\documentclass[11pt,a4paper]{article}
\usepackage[hyperref]{eacl2021}
\usepackage{times}
\usepackage{latexsym}

\usepackage{microtype}

\aclfinalcopy 
\def\aclpaperid{***} 

\title{\texttt{BERTić} - The Transformer Language Model for \\Bosnian, Croatian, Montenegrin and Serbian}

\author{Nikola Ljubešić \\
  Jožef Stefan Institute \\
  Jamova cesta 39\\
  Ljubljana, Slovenia \\
  \texttt{nikola.ljubesic@ijs.si} \\\And
  Davor Lauc \\
  Faculty of Humanities and Social Sciences \\
  Ivana Lučića 3\\
  Zagreb, Croatia \\
  \texttt{davor.lauc@ffzg.hr} \\}

\date{}

\begin{document}
\maketitle
\begin{abstract}
In this paper we describe a transformer model pre-trained on 8 billion tokens of crawled text from the Croatian, Bosnian, Serbian and Montenegrin web domains. We evaluate the transformer model on the tasks of part-of-speech tagging, named-entity-recognition, geo-location prediction and commonsense causal reasoning, showing improvements on all tasks over state-of-the-art models. For commonsense reasoning evaluation we introduce COPA-HR - a translation of the Choice of Plausible Alternatives (COPA) dataset into Croatian. The \texttt{BERTić} model is made available for free usage and further task-specific fine-tuning through HuggingFace.
\end{abstract}

\section{Introduction}

In recent years, pre-trained transformer models have taken the NLP world by storm~\cite{devlin2018bert, liu2019roberta, brown2020language}, yielding new state-of-the-art results in various tasks and settings. While such models, requiring significant computing power and data quantity, started to emerge for non-English languages~\cite{martin2019camembert, de2019bertje}, as well as in multilingual flavours~\cite{devlin2018bert,conneau2019unsupervised}, there is a significant number of languages for which better models can be obtained with the available pre-training techniques.

This paper describes such an effort - training a transformer language model on more than 8 billion tokens of text written in the Bosnian, Croatian, Montenegrin or Serbian language, all these languages being very closely related, mutually intelligible, and classified under the same HBS (Serbo-Croatian) macro-language by the ISO-693-3 standard.\footnote{\url{https://iso639-3.sil.org/code_tables/macrolanguage_mappings/data}}

The name of the model -- \texttt{BERTić} -- points at two facts: (1) the language model was trained in Zagreb, Croatia, in whose vernacular diminutives ending in \emph{ić} are frequently used (\emph{fotić} eng. photo camera, \emph{smajlić} eng. smiley, \emph{hengić} eng. hanging together), and (2) in all the countries / languages of this model the patronymic surnames end to a great part with the suffix \emph{ić}, with likely diminutive etymology.

The paper is structured as follows: in the following section we describe the data the model is based on, in the third section we give a short description of the modelling performed, and in the fourth section we present a detailed evaluation of the model.

\section{Data}

As our data basis we use already existing datasets, namely (1) the hrWaC corpus of the Croatian top-level domain, crawled in 2011~\cite{ljubevsic2011hrwac} and again in 2014~\cite{ljubevsic2014bs}, (2) the srWaC corpus of the Serbian top-level domain, crawled in 2014~\cite{ljubevsic2014bs}, (3) the bsWaC corpus of the Bosnian top-level domain, crawled in 2014~\cite{ljubevsic2014bs}, (4) the cnrWaC corpus of the Montenegrin top-level domain, crawled in 2019, and (5) the Riznica corpus consisting of Croatian literary works and newspapers~\cite{cavar2012riznica}.

Given that most of the crawls contain data only up to year 2014, we performed new crawls of the Bosnian, Croatian and Serbian top-level domains. We brand these corpora as CLASSLA web corpora given that CLASSLA is the CLARIN knowledge centre for South Slavic languages\footnote{\url{https://www.clarin.si/info/k-centre/}} under which we perform most of the described activities. We deduplicate the CLASSLA corpora by removing identical sentences that were already present in the WaC corpora. 
The amount of data removed through this deduplication is minor, in all cases in single digit percentages.

We further exploit the recently published cc100 corpora~\cite{conneau2019unsupervised} that are based on the CommonCrawl data collection. We perform the same level of deduplication as with the CLASSLA corpora, with every sentence already present in the WaC or CLASSLA corpus being removed from the cc100 corpus. 
This round of deduplication removed around 15\% of the CommonCrawl data.




The resulting sizes of the datasets used for training the \texttt{BERTić} model are presented in Table \ref{tab:data}. The overall text collection consists of 8,387,681,518 words.

\begin{table}
    \centering
\begin{tabular}{l|l|r}
dataset & language & \# of words \\
\hline
hrWaC & Croatian & 1,250,923,836 \\
CLASSLA-hr & Croatian & 1,341,494,461 \\
cc100-hr & Croatian &  2,880,490,449 \\
Riznica & Croatian & 87,724,737 \\
srWaC & Serbian & 493,202,149 \\
CLASSLA-sr & Serbian & 752,916,260 \\
cc100-sr & Serbian & 711,014,370 \\
bsWaC & Bosnian & 256,388,597 \\
CLASSLA-bs & Bosnian & 534,074,921 \\
cnrWaC & Montenegrin & 79,451,738 \\
\end{tabular}
\caption{\label{tab:data}Datasets used for training the \texttt{BERTić} model with their size (in number of words) after deduplication.}
\end{table}

\section{Model training}

For training this model we selected the Electra approach to training transformer models~\cite{clark2020electra}. These models are based on training a smaller generator model and the main, larger, discriminator model whose task is to discriminate whether a specific word is the original word from the text, or a word generated by the generator model. The authors claim that the Electra approach is computationally more efficient than the BERT models~\cite{devlin2018bert} based on masked language modelling.

As in BERT and similar transformers models, we constructed a WordPiece vocabulary with a vocabulary size of 32 thousand tokens. A WordPiece model was trained using the HuggingFace tokenizers library\footnote{\url{https://huggingface.co/transformers/main_classes/tokenizer.html}} on the random sample of 10 million paragraphs from the whole dataset. Text pre-processing and cleaning differ from the original BERT only in preserving all Unicode characters, while in the original pre-processing diacritics are removed.

Training of the model was performed to the most part with the hyperparameters set for base-sized models (110 million parameters in 12 transformer layers) as defined in the Electra paper~\cite{clark2020electra}. Training batch size was kept at 1024, the maximum size for the 8 TPUv3 units on which the training was performed. The training was run for 2 million steps (roughly 50 epochs).

\section{Evaluation}

In this section we present an exhaustive evaluation of the newly trained \texttt{BERTić} model on two token classification tasks -- morphosyntactic tagging and named entity recognition, and two sequence classification tasks -- geolocation prediction and commonsense causative reasoning.

The reference points in each task are the state-of-the art transformer models covering the macro-language - multilingual BERT~\cite{devlin2018bert} and CroSloEngual BERT~\cite{ulvcar2020finest}. While multilingual BERT (\texttt{mBERT} onwards) was trained on Wikipedia corpora, CroSloEngual BERT (\texttt{cseBERT} onwards) was trained on a similar amount of Croatian data used to train \texttt{BERTić}, but without the data from the remaining languages.

\begin{table*}
\begin{center}
\begin{tabular}{l|l|l|rrrr}
dataset & language & variety & \texttt{CLASSLA} & \texttt{mBERT} & \texttt{cseBERT} & \texttt{BERTić} \\
\hline
hr500k & Croatian & standard & 93.87 & 94.60 & 95.74 & \textbf{***95.81} \\
reldi-hr & Croatian & non-standard & - & 88.87 & 91.63 & \textbf{***92.28} \\
SETimes.SR & Serbian & standard & 95.00 & 95.50 & \textbf{96.41} & 96.31 \\
reldi-sr & Serbian & non-standard & - & 91.26 & 93.54 & \textbf{***93.90}\\
\end{tabular}
\end{center}
\caption{\label{tab:pos}Average microF1 results on the morphosyntactic annotation task over five training iterations. The highest score per dataset is marked with bold. The statistical significance is tested with the two-sided t-test over the five runs between the two strongest results. Level of significance is labeled with asteriks signs (*** p$<=$0.001).}
\end{table*}

\begin{table*}
\begin{center}
\begin{tabular}{l|l|l|rrrr}
dataset & language & variety & \texttt{CLASSLA} & \texttt{mBERT} & \texttt{cseBERT} & \texttt{BERTić} \\
\hline
hr500k & Croatian & standard & 80.13 & 85.67 & 88.98 & \textbf{****89.21}\\
ReLDI-hr & Croatian & non-standard & - & 76.06 & 81.38 & \textbf{****83.05} \\
SETimes.SR & Serbian & standard & 84.64 & \textbf{92.41} & 92.28 & 92.02 \\
ReLDI-sr & Serbian & non-standard &  - & 81.29 & 82.76 & \textbf{***87.92}\\
\end{tabular}
\end{center}
\caption{\label{tab:ner}Average F1 results on the named entity recognition task over five training iterations. The highest score per dataset is marked with bold. The statistical significance is tested with the two-sided t-test over the five runs between the two strongest results. Level of significance is labeled with asteriks signs (*** p$<=$0.001, **** p$<=$0.0001).}
\end{table*}

\subsection{Morphosyntactic tagging\label{sec:pos}}

On the task of morphosyntactic tagging (assigning each word one among multiple hundreds of detailed morphosyntactic classes, e.g. \texttt{Ncmsay} referring to a common masculine noun, in accusative case, singular number, animate) we compare the three transformer models, \texttt{mBERT}, \texttt{cseBERT} and \texttt{BERTić}. We additionally report results, when available, for the current production tagger for the two languages - the \texttt{CLASSLA} tool~\cite{ljubesic-dobrovoljc-2019-neural}, based on Stanford's Stanza, exploiting static embedding and BiLSTM technology~\cite{qi2020stanza}.

We perform evaluation of the models on this task on four datasets: the Croatian standard language dataset hr500k~\cite{11356/1183}, the Croatian non-standard language dataset ReLDI-hr~\cite{11356/1241}, the Serbian standard language dataset SETimes.SR~\cite{11356/1200} and the Serbian non-standard Twitter language dataset ReLDI-sr~\cite{11356/1240}.

For each dataset and model we perform hyperparameter optimization via Bayesian search on the wandb.ai platform~\cite{wandb}, allowing for 30 iterations. We optimize the initial learning rate (we search between the values of 9e-6 and 1e-4) and the epoch number (we search between the values of 3 and 15).

We report average microF1 results of five runs per dataset and model in Table \ref{tab:pos}. The highest score per dataset is marked with bold. The statistical significance is tested with the two-sided t-test over the five runs between the two highest average results. We can observe that the \texttt{BERTić} model outperforms all the remaining models, \texttt{cseBERT} coming second, on three out of four datasets. Only on the Serbian standard dataset the difference between these two models is insignificant. We argue that this is due to the simplicity of the dataset - it consists of texts from one newspaper only, therefore containing text with little variation even between the training and the testing data.

\subsection{Named entity recognition}

On the task of named entity recognition we compare the same models on the same datasets as was the case in the previous Section \ref{sec:pos}. We also perform an identical hyperparameter optimisation and experimentation and report the results in Table \ref{tab:ner}. The results show again that the two best performing models are \texttt{cseBERT} and \texttt{BERTić} with the latter performing better on three out of four datasets, again, with no significant difference on the standard Serbian task for the same reasons as with the previous task.

\subsection{Social media geolocation}

In this subsection we compare the three transformer models on the Social Media Geolocation (SMG2020) shared task, which part of the VarDial 2020 Evaluation Campaign~\cite{gaman2020report}. The task consists of predicting the exact latitude and longitude of a geo-encoded tweet published in Croatia, Bosnia, Montenegro or Serbia. The shared task winner in 2020 was using the \texttt{cseBERT} model~\cite{scherrer2020helju} in its approach.

We evaluate the model on the two evaluation metrics of the shared task - median and mean of the distance between gold and predicted geolocations. Given the large size of the training dataset (320,042 instances), we do not perform any additional hyperparameter tuning beyond the one performed during the participation in the shared task and apply the same methodology: we fine-tune the transformer model with batch size of 64 for 40 epochs and retain the model with minimum median distance on development data. The results in Table \ref{tab:smg} show that the \texttt{BERTić} model improved the results of the shared task winner -- the \texttt{cseBERT} model.

\begin{table}
\begin{center}
\begin{tabular}{l|rr}
& median & mean \\
\hline
centroid & 107.10 & 145.72 \\
\texttt{mBERT} & 42.25 & 82.05 \\
\texttt{cseBERT} & 40.76 & 81.88 \\ 
\texttt{BERTić} & \textbf{37.96} & \textbf{79.30} \\
\end{tabular}
\end{center}
\caption{\label{tab:smg} Median distance and mean distance between gold and predicted geolocation (lower is better) on the task of social media geolocation prediction. The best results are marked in bold. No statistical testing was performed due to a large size of the test dataset (39,723 instances).}
\end{table}

\subsection{Commonsense causal reasoning}

The final evaluation round of the new \texttt{BERTić} model is performed on the task of commonsense causal reasoning on a translation of the COPA dataset~\cite{roemmele2011choice} into Croatian, the COPA-HR dataset. The translation is performed by following the methodology laid out while preparing the XCOPA dataset~\cite{ponti-etal-2020-xcopa}, a translation of the COPA dataset into 11 typologically balanced languages.

The dataset consists of 400 training, 100 development and 500 examples. 
Each instance in the dataset consists of a premise (\emph{The man broke his toe}), a question (\emph{What was the cause?}),\footnote{Roughly half of the instances contain the other question: \emph{What was the effect?},} and two alternatives, one of them to be chosen by the system as being more plausible (\emph{He got a hole in his sock}, or \emph{He dropped a hammer on his foot}).

While translating the dataset, the translator was also given the task of selecting the more plausible alternative given their translation. The observed agreement between the annotations in the English dataset and the annotations of the Croatian translator was perfect on the training set and the development set, while on the test set one out of 500 choices differed. The problematic example proved to be a rather unclear case -- the premise being \emph{I paused to stop talking.}, with the question \emph{What was the cause?}, and the alternatives \emph{I lost my voice.} and \emph{I ran out of breath.}.\footnote{The Croatian translator chose the second alternative, while in the original dataset the first alternative is chosen.} The dataset is available from the CLARIN.SI repository~\cite{11356/1404}.\footnote{\url{http://hdl.handle.net/11356/1404}}

The approach taken to benchmarking the three transformer models is that of sentence pair classification, each original instance becoming two sentence pair instances (each sentence pair containing the premise and one alternative), with different models being trained for \emph{cause} and \emph{effect} questions. 
During evaluation, separate predictions are made on each of the alternatives, the per-class predictions being fed to a softmax function, and the higher positive-class alternative being chosen as the correct one.

The standard evaluation metric for this dataset is accuracy. Given the balanced nature of the test set, the random baseline is 50\%. For hyperparameter optimization the same approach was taken as with the token classification tasks.

The results presented in Table \ref{tab:copa} show that both language-specific transformer models outperform \texttt{mBERT} significantly, with \texttt{BERTić} obtaining a significant lead over \texttt{cseBERT}.

\begin{table}
\begin{center}
\begin{tabular}{l|r}
& accuracy \\
\hline
random & 50.00 \\
\texttt{mBERT} & 54.12 \\
\texttt{cseBERT} & 61.80 \\
\texttt{BERTić} & \textbf{**65.76} \\
\end{tabular}
\end{center}
\caption{\label{tab:copa}Average accuracy results on the commonsense causal reasoning task over five training iterations. The highest score per dataset is marked with bold. The statistical significance is tested with the two-sided t-test over the five runs between the two strongest results (** p$<=$0.01).}
\end{table}

\section{Conclusion}

In this paper we have presented a newly published Electra transformer language model, \texttt{BERTić}, trained on more than 8 billion tokens of previously and newly collected web text written in Bosnian, Croatian, Montenegrin or Serbian. We have applied a very thorough evaluation of the model, comparing it primarily to other state-of-the-art transformer models that support the languages in question. We have obtained significant improvements on all four tasks, with no difference obtained only on one single-source-dataset with little text variation and high training and testing data similarity.

The main conclusions we can draw from our results are the following. (1) Although \texttt{cseBERT} and \texttt{BERTić} use a different approach to building transformer language models, our assumption is that the performance difference between these two lies primarily in the larger amount of data presented to the \texttt{BERTić} model. (2) The improvements on the four tasks with the \texttt{BERTić} model seem to be smaller on the morphosyntactic tagging task than the remaining three tasks that require more world and commonsense reasoning knowledge. (3) Except for the named entity recognition task on the Serbian non-standard dataset, we fail to observe greater improvements on Serbian tasks than on Croatian ones between \texttt{cseBERT} and \texttt{BERTić}, regardless the fact that the former has seen none and the latter has seen huge quantities of Serbian text, showing the irrelevance of minor language differences for performance of large transformer models. (4) While BiLSTM models are still close-to-competitive on the morphosyntactic tagging task, they cannot hold up on the named entity recognition task as it requires more common knowledge. Such knowledge transformer models manage to absorb to a much higher level than pre-trained static embeddings used by BiLSTMs.

The \texttt{BERTić} model is available from the HuggingFace repository at \url{https://huggingface.co/classla/bcms-bertic}.

\section*{Acknowledgments}

This work has been supported by the Slovenian Research Agency and the Flemish Research Foundation through the bilateral research project ARRS N6-0099 and FWO G070619N ``The linguistic landscape of hate speech on social media'', the Slovenian Research Agency research core funding No. P6-0411 ``Language resources and technologies for Slovene language'', and the European Union’s Rights, Equality and Citizenship Programme (2014-2020) project IMSyPP (grant no. 875263). We would like to thank the anonymous reviewers and Ivo-Pavao Jazbec for their useful feedback.

\bibliography{eacl2021}

\begin{thebibliography}{23}
\expandafter\ifx\csname natexlab\endcsname\relax\def\natexlab#1{#1}\fi

\bibitem[{Batanovi{\'c} et~al.(2018)Batanovi{\'c}, Ljube{\v s}i{\'c}, Samard{\v
  z}i{\'c}, and Erjavec}]{11356/1200}
Vuk Batanovi{\'c}, Nikola Ljube{\v s}i{\'c}, Tanja Samard{\v z}i{\'c}, and
  Toma{\v z} Erjavec. 2018.
\newblock \href {http://hdl.handle.net/11356/1200} {Training corpus
  {SETimes}.{SR} 1.0}.
\newblock Slovenian language resource repository {CLARIN}.{SI}.

\bibitem[{Biewald(2020)}]{wandb}
Lukas Biewald. 2020.
\newblock \href {https://www.wandb.com/} {Experiment tracking with weights and
  biases}.
\newblock Software available from wandb.com.

\bibitem[{Brown et~al.(2020)Brown, Mann, Ryder, Subbiah, Kaplan, Dhariwal,
  Neelakantan, Shyam, Sastry, Askell et~al.}]{brown2020language}
Tom~B Brown, Benjamin Mann, Nick Ryder, Melanie Subbiah, Jared Kaplan, Prafulla
  Dhariwal, Arvind Neelakantan, Pranav Shyam, Girish Sastry, Amanda Askell,
  et~al. 2020.
\newblock Language models are few-shot learners.
\newblock \emph{arXiv preprint arXiv:2005.14165}.

\bibitem[{{\'C}avar and Ron{\v{c}}evi{\'c}(2012)}]{cavar2012riznica}
Damir {\'C}avar and Dunja~Brozovi{\'c} Ron{\v{c}}evi{\'c}. 2012.
\newblock {Riznica: the Croatian language corpus}.
\newblock \emph{Prace filologiczne}, 63:51--65.

\bibitem[{Clark et~al.(2020)Clark, Luong, Le, and Manning}]{clark2020electra}
Kevin Clark, Minh-Thang Luong, Quoc~V Le, and Christopher~D Manning. 2020.
\newblock Electra: Pre-training text encoders as discriminators rather than
  generators.
\newblock \emph{arXiv preprint arXiv:2003.10555}.

\bibitem[{Conneau et~al.(2019)Conneau, Khandelwal, Goyal, Chaudhary, Wenzek,
  Guzm{\'a}n, Grave, Ott, Zettlemoyer, and Stoyanov}]{conneau2019unsupervised}
Alexis Conneau, Kartikay Khandelwal, Naman Goyal, Vishrav Chaudhary, Guillaume
  Wenzek, Francisco Guzm{\'a}n, Edouard Grave, Myle Ott, Luke Zettlemoyer, and
  Veselin Stoyanov. 2019.
\newblock Unsupervised cross-lingual representation learning at scale.
\newblock \emph{arXiv preprint arXiv:1911.02116}.

\bibitem[{Devlin et~al.(2018)Devlin, Chang, Lee, and
  Toutanova}]{devlin2018bert}
Jacob Devlin, Ming-Wei Chang, Kenton Lee, and Kristina Toutanova. 2018.
\newblock {Bert: Pre-training of deep bidirectional transformers for language
  understanding}.
\newblock \emph{arXiv preprint arXiv:1810.04805}.

\bibitem[{Gaman et~al.(2020)Gaman, Hovy, Ionescu, Jauhiainen, Jauhiainen,
  Lind{\'e}n, Ljube{\v{s}}i{\'c}, Partanen, Purschke, Scherrer
  et~al.}]{gaman2020report}
Mihaela Gaman, Dirk Hovy, Radu~Tudor Ionescu, Heidi Jauhiainen, Tommi
  Jauhiainen, Krister Lind{\'e}n, Nikola Ljube{\v{s}}i{\'c}, Niko Partanen,
  Christoph Purschke, Yves Scherrer, et~al. 2020.
\newblock {A report on the VarDial evaluation campaign 2020}.
\newblock In \emph{Proceedings of the 7th Workshop on NLP for Similar
  Languages, Varieties and Dialects}. International Committee on Computational
  Linguistics.

\bibitem[{Liu et~al.(2019)Liu, Ott, Goyal, Du, Joshi, Chen, Levy, Lewis,
  Zettlemoyer, and Stoyanov}]{liu2019roberta}
Yinhan Liu, Myle Ott, Naman Goyal, Jingfei Du, Mandar Joshi, Danqi Chen, Omer
  Levy, Mike Lewis, Luke Zettlemoyer, and Veselin Stoyanov. 2019.
\newblock {Roberta: A robustly optimized bert pretraining approach}.
\newblock \emph{arXiv preprint arXiv:1907.11692}.

\bibitem[{Ljube{\v s}i{\'c}(2021)}]{11356/1404}
Nikola Ljube{\v s}i{\'c}. 2021.
\newblock \href {http://hdl.handle.net/11356/1404} {Choice of plausible
  alternatives dataset in {C}roatian {COPA}-{HR}}.
\newblock Slovenian language resource repository {CLARIN}.{SI}.

\bibitem[{Ljube{\v s}i{\'c} et~al.(2018)Ljube{\v s}i{\'c}, Agi{\'c}, Klubi{\v
  c}ka, Batanovi{\'c}, and Erjavec}]{11356/1183}
Nikola Ljube{\v s}i{\'c}, {\v Z}eljko Agi{\'c}, Filip Klubi{\v c}ka, Vuk
  Batanovi{\'c}, and Toma{\v z} Erjavec. 2018.
\newblock \href {http://hdl.handle.net/11356/1183} {Training corpus hr500k
  1.0}.
\newblock Slovenian language resource repository {CLARIN}.{SI}.

\bibitem[{Ljube{\v{s}}i{\'c} and
  Dobrovoljc(2019)}]{ljubesic-dobrovoljc-2019-neural}
Nikola Ljube{\v{s}}i{\'c} and Kaja Dobrovoljc. 2019.
\newblock \href {https://doi.org/10.18653/v1/W19-3704} {What does neural bring?
  analysing improvements in morphosyntactic annotation and lemmatisation of
  {S}lovenian, {C}roatian and {S}erbian}.
\newblock In \emph{Proceedings of the 7th Workshop on Balto-Slavic Natural
  Language Processing}, pages 29--34, Florence, Italy. Association for
  Computational Linguistics.

\bibitem[{Ljube{\v{s}}i{\'c} and Erjavec(2011)}]{ljubevsic2011hrwac}
Nikola Ljube{\v{s}}i{\'c} and Toma{\v{z}} Erjavec. 2011.
\newblock {hrWaC and slWaC: Compiling web corpora for Croatian and Slovene}.
\newblock In \emph{International Conference on Text, Speech and Dialogue},
  pages 395--402. Springer.

\bibitem[{Ljube{\v s}i{\'c} et~al.(2019{\natexlab{a}})Ljube{\v s}i{\'c},
  Erjavec, Batanovi{\'c}, Mili{\v c}evi{\'c}, and Samard{\v
  z}i{\'c}}]{11356/1241}
Nikola Ljube{\v s}i{\'c}, Toma{\v z} Erjavec, Vuk Batanovi{\'c}, Maja Mili{\v
  c}evi{\'c}, and Tanja Samard{\v z}i{\'c}. 2019{\natexlab{a}}.
\newblock \href {http://hdl.handle.net/11356/1241} {Croatian twitter training
  corpus {ReLDI}-{NormTagNER}-hr 2.1}.
\newblock Slovenian language resource repository {CLARIN}.{SI}.

\bibitem[{Ljube{\v s}i{\'c} et~al.(2019{\natexlab{b}})Ljube{\v s}i{\'c},
  Erjavec, Batanovi{\'c}, Mili{\v c}evi{\'c}, and Samard{\v
  z}i{\'c}}]{11356/1240}
Nikola Ljube{\v s}i{\'c}, Toma{\v z} Erjavec, Vuk Batanovi{\'c}, Maja Mili{\v
  c}evi{\'c}, and Tanja Samard{\v z}i{\'c}. 2019{\natexlab{b}}.
\newblock \href {http://hdl.handle.net/11356/1240} {Serbian twitter training
  corpus {ReLDI}-{NormTagNER}-sr 2.1}.
\newblock Slovenian language resource repository {CLARIN}.{SI}.

\bibitem[{Ljube{\v{s}}i{\'c} and Klubi{\v{c}}ka(2014)}]{ljubevsic2014bs}
Nikola Ljube{\v{s}}i{\'c} and Filip Klubi{\v{c}}ka. 2014.
\newblock {$\{$bs, hr, sr$\}$ wac-web corpora of Bosnian, Croatian and
  Serbian}.
\newblock In \emph{Proceedings of the 9th Web as Corpus Workshop (WaC-9)},
  pages 29--35.

\bibitem[{Martin et~al.(2019)Martin, Muller, Su{\'a}rez, Dupont, Romary, de~la
  Clergerie, Seddah, and Sagot}]{martin2019camembert}
Louis Martin, Benjamin Muller, Pedro Javier~Ortiz Su{\'a}rez, Yoann Dupont,
  Laurent Romary, {\'E}ric~Villemonte de~la Clergerie, Djam{\'e} Seddah, and
  Beno{\^\i}t Sagot. 2019.
\newblock {Camembert: a tasty french language model}.
\newblock \emph{arXiv preprint arXiv:1911.03894}.

\bibitem[{Ponti et~al.(2020)Ponti, Glava{\v{s}}, Majewska, Liu, Vuli{\'c}, and
  Korhonen}]{ponti-etal-2020-xcopa}
Edoardo~Maria Ponti, Goran Glava{\v{s}}, Olga Majewska, Qianchu Liu, Ivan
  Vuli{\'c}, and Anna Korhonen. 2020.
\newblock \href {https://doi.org/10.18653/v1/2020.emnlp-main.185} {{XCOPA}: A
  multilingual dataset for causal commonsense reasoning}.
\newblock In \emph{Proceedings of the 2020 Conference on Empirical Methods in
  Natural Language Processing (EMNLP)}, pages 2362--2376, Online. Association
  for Computational Linguistics.

\bibitem[{Qi et~al.(2020)Qi, Zhang, Zhang, Bolton, and Manning}]{qi2020stanza}
Peng Qi, Yuhao Zhang, Yuhui Zhang, Jason Bolton, and Christopher~D Manning.
  2020.
\newblock {Stanza: A Python natural language processing toolkit for many human
  languages}.
\newblock \emph{arXiv preprint arXiv:2003.07082}.

\bibitem[{Roemmele et~al.(2011)Roemmele, Bejan, and
  Gordon}]{roemmele2011choice}
Melissa Roemmele, Cosmin~Adrian Bejan, and Andrew~S Gordon. 2011.
\newblock {Choice of Plausible Alternatives: An Evaluation of Commonsense
  Causal Reasoning.}
\newblock In \emph{AAAI Spring Symposium: Logical Formalizations of Commonsense
  Reasoning}, pages 90--95.

\bibitem[{Scherrer and Ljube{\v{s}}i{\'c}(2020)}]{scherrer2020helju}
Yves Scherrer and Nikola Ljube{\v{s}}i{\'c}. 2020.
\newblock {HeLju@ VarDial 2020: Social media variety geolocation with BERT
  models}.
\newblock In \emph{Proceedings of the 7th Workshop on NLP for Similar
  Languages, Varieties and Dialects}, pages 202--211.

\bibitem[{Ul{\v{c}}ar and Robnik-{\v{S}}ikonja(2020)}]{ulvcar2020finest}
Matej Ul{\v{c}}ar and Marko Robnik-{\v{S}}ikonja. 2020.
\newblock {FinEst BERT and CroSloEngual BERT}.
\newblock In \emph{International Conference on Text, Speech, and Dialogue},
  pages 104--111. Springer.

\bibitem[{de~Vries et~al.(2019)de~Vries, van Cranenburgh, Bisazza, Caselli, van
  Noord, and Nissim}]{de2019bertje}
Wietse de~Vries, Andreas van Cranenburgh, Arianna Bisazza, Tommaso Caselli,
  Gertjan van Noord, and Malvina Nissim. 2019.
\newblock {Bertje: A dutch BERT model}.
\newblock \emph{arXiv preprint arXiv:1912.09582}.

\end{thebibliography}
\bibliographystyle{acl_natbib}

\end{document}